\documentclass[10pt,twocolumn,letterpaper]{article}

\usepackage{cvpr}      

\usepackage{graphicx}
\usepackage{amsmath}
\usepackage{amssymb}
\usepackage{booktabs}
\usepackage[accsupp]{axessibility}  

\usepackage[pagebackref,breaklinks,colorlinks]{hyperref}

\usepackage[capitalize]{cleveref}
\crefname{section}{Sec.}{Secs.}
\Crefname{section}{Section}{Sections}
\Crefname{table}{Table}{Tables}
\crefname{table}{Tab.}{Tabs.}

\def\cvprPaperID{17} 
\def\confName{CVPR}
\def\confYear{2023}

\begin{document}

\title{SUPRA: Superpixel Guided Loss for Improved Multi-modal \\ Segmentation in Endoscopy}

\author{Rafael Martinez-Garcia-Peña, Mansoor Ali Teevno, Gilberto Ochoa-Ruiz*\\
Tecnologico de Monterrey, School of Sciences and Engineering\\
Av. Eugenio Garza Sada 2501 Sur, Tecnológico, 64849 Monterrey, N.L.\\
{\tt\small *Corresponding: gilberto.ochoa@tec.mx}
\and
Sharib Ali*\\
University of Leeds, Computing Department\\
Woodhouse, Leeds LS2 9JT, United Kingdom\\
{\tt\small *Corresponding: S.S.Ali@leeds.ac.uk}
}
\maketitle

\begin{abstract}
Domain shift is a well-known problem in the medical imaging community. In particular, for endoscopic image analysis data can have different modalities that cause the performance of deep learning (DL) methods to become adversely affected. Methods developed on one modality cannot be used for a different modality without retraining. However, in real clinical settings, endoscopists switch between modalities depending on the specifics of the condition being explored. In this paper, we explore domain generalisation to enable DL methods to be used in such scenarios. To this extent, we propose to use superpixels generated with Simple Linear Iterative Clustering (SLIC) which we refer to as ``SUPRA'' for SUPeRpixel Augmented method. SUPRA first generates a preliminary segmentation mask making use of our new loss  ``SLICLoss'' that encourages both an accurate and superpixel-consistent segmentation. We demonstrate that SLICLoss when combined with Binary Cross Entropy loss (BCE) can improve the model's generalisability with data that presents significant domain shift due to a change in lighting modalities. We validate this novel compound loss on a vanilla UNet using the EndoUDA dataset, which contains images for Barret's Esophagus from two modalities. We show that our method yields a relative improvement of more than 20\% IoU in the target domain set compared to the baseline.

\end{abstract}

\section{Introduction}
Computer vision has traditionally been an area of significant interest in many practical tasks. In biomedical and surgical fields, imaging and image analysis techniques shown potential for various automated tasks such as segmentation \cite{pone}, tracking, and detection in clinical settings \cite{ali_comprehensive_2022}. Furthermore, deep learning-based (DL) methods are increasingly being developed and deployed in biomedical imaging -- notably in endoscopies \cite{npj_DM2023} -- as data becomes more available and procedures become more complex \cite{luo_advanced_2018}.

\begin{figure}
\center
  \includegraphics[width=0.38\textwidth]{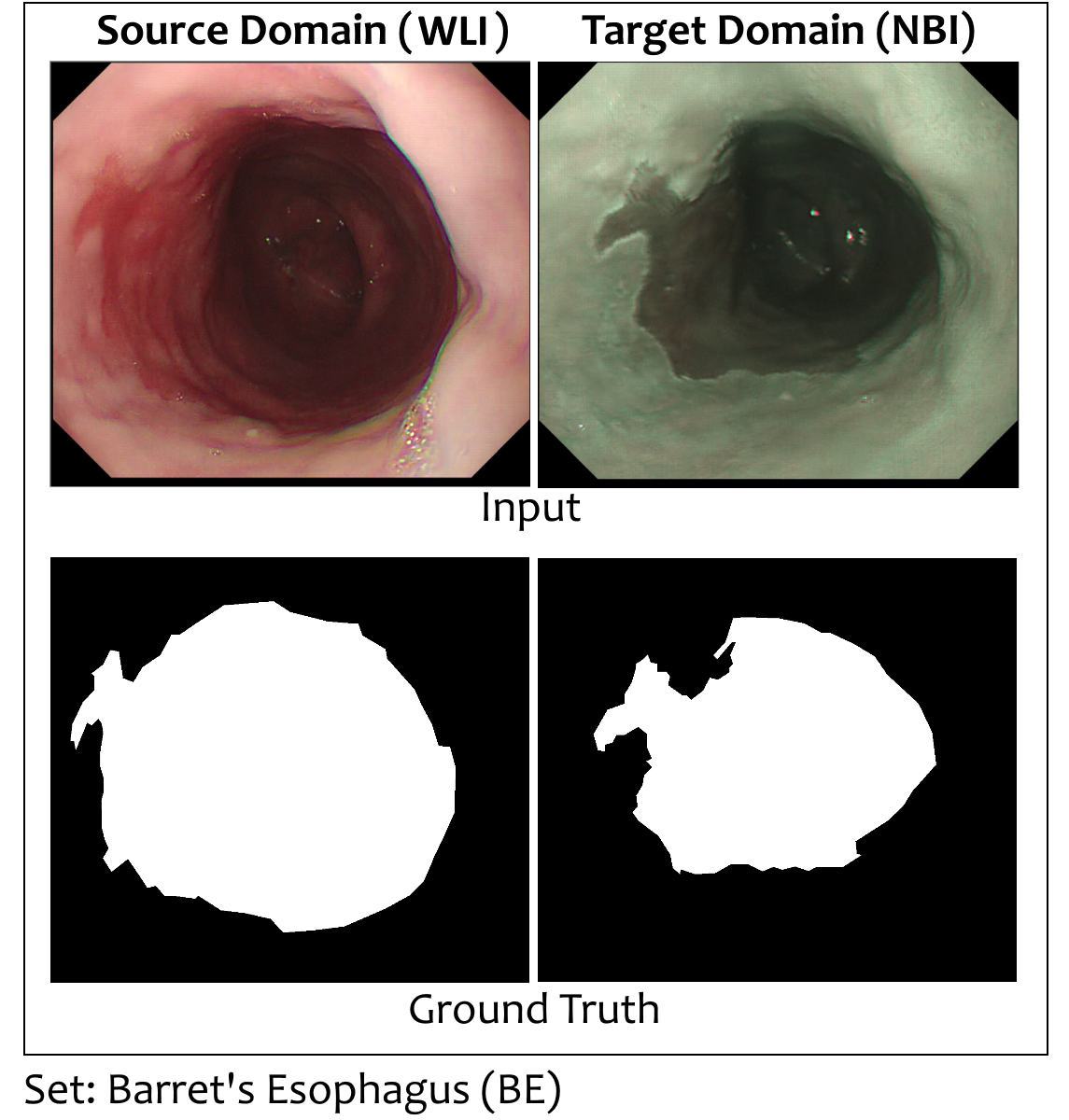}
  \caption{\textbf{Sample images from EndoUDA.} These show the widely used modality white light imaging (WLI, left) and a narrow-band imaging frame (NBI, right) for Barret's Esophagus (BE)\cite{celik_endouda_2021}.}
  \label{fig:endouda}
\end{figure}

\begin{figure*}[htbp]
\centerline{\includegraphics[width=1\textwidth]{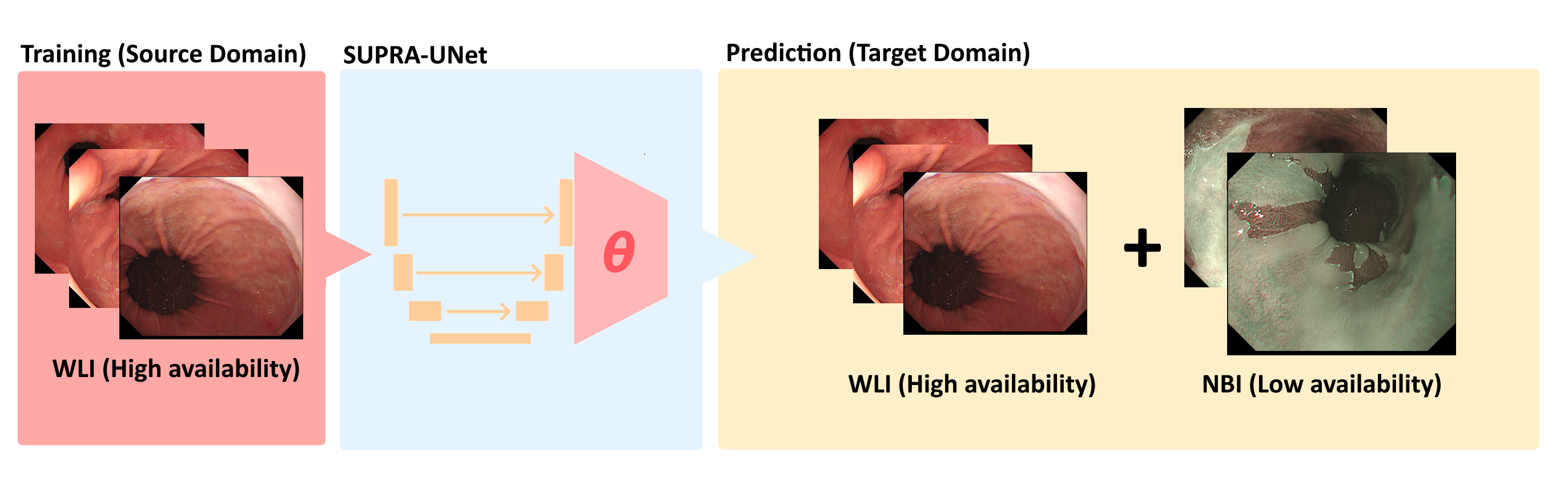}}
\caption{\textbf{A schematized view of our problem}. We use frames with high availability (easy to acquire both frames and labels in a sufficient volume for training) as our source domain for training. We use white-light imaging (WLI) for this purpose, as it is widely used for endoscopic surveillance. Our model is trained using only data from WLI (source) modality, requiring no extra information about any other target modalities (in our case, Narrow-band imaging, NBI). During the prediction phase our model is expected to perform well not only for WLI, but also for the target NBI modality. NBI is chosen here due to the lack of expert level annotations available. We present a domain generalization technique that is able to tackle this problem using a superpixel augmented network referred to as 'SUPRA-UNet'.
}
\label{fig:problem}
\end{figure*}

These advances, however, have also highlighted several shortcomings in many DL methods, which arise from the nature of the medical setting: not only are the tasks themselves quite difficult, but data availability is a significant hurdle for most of the common methods used successfully in other domains \cite{ros_comparative_2021}. Furthermore, when there is available data, it tends to be limited to only a small fraction of the cases one would observe in real-world scenarios: everything from instruments and acquisition devices, to procedures used  and the lighting conditions are variable and can change from a hospital-to-hospital \cite{jc-tec-2022}.

One of such tasks where these data constraints become apparent is in computer vision applications for endoscopy. Acquisition of medical data is expensive, and time consuming \cite{liu_unsupervised_2020}. Obtaining enough volume to improve training in specific tasks while accounting for variations introduced by different patients, instruments, and the complex nature of the environment is not a trivial task. To complicate matters further, there may be task-specific subdomains that use instrumentation that can significantly alter the visual properties of the frames being captured, limiting the usefulness of models not trained to perform in that subdomain.

An example of this limitation is the usage of different light modalities during an endoscopic examination. White-Light Imaging (WLI) may be used for a general examination, with more specific areas highlighted with Narrow-Band Imaging (NBI), allowing a clinician to inspect different anatomical aspects of the same lesion \cite{barbeiro_narrow-band_2019, kato_magnifying_2010}. This is the case for a condition known as Barrett's Esophagus (BE), which involves the presence of columnar epithelium (the type of lining that usually covers the stomach and intestines) in the esophagus instead of the squamous epithelium that is normal in this area of the body \cite{spechler_barretts_1986}. By itself, this condition is not dangerous, but the presence of this altered epithelium is correlated with heightened risk of esophageal cancer. Early detection is important, as BE is asymptomatic aside from its relation to Gastroesophageal Reflux Disease (GERD, colloquially acid reflux)\cite{clarrett_gastroesophageal_2018, falk_barretts_2002}.

This makes BE interesting for image segmentation for two reasons: First, early and accurate detection during routine endoscopic examinations is critical to reduce the risk of more dangerous diseases \cite{evans_role_2012}. Second, the use of different lighting modalities can improve the ability of a clinician to detect and inspect the sites where the condition could be present -- particularly with the use of NBI to inspect the boundary between squamous and columnar epithelium \cite{kato_magnifying_2010}. 

If a model for endoscopic computer aided diagnostic tools is to be usable in this setting, then it is important that such models can work seamlessly in both of these imaging modalities, and that they avoid the need for any modality-specific training. As shown in fig.  \ref{fig:endouda} (which shows two sample images from the EndoUDA dataset using two different modalities) the changes introduced by switching the lighting modality strongly alter the visual properties of the areas to be segmented \cite{celik_endouda_2021}. These changes affect a model's ability to generalise. This problem is further exacerbated due to the relatively limited amount of NBI images compared to those available for WLI.

If one uses more of the available data for WLI, then we must envisage methods that can reduce the impact of the domain shift when the Source Domain (SD) is WLI and the Target Domain (TD) is NBI. The problem is described graphically in fig. \ref{fig:problem}. Developing such a modality agnostic approach for the segmentation of injuries in the esophagus is the focus of this work. \\

\begin{figure*}[htbp]
\centerline{\includegraphics[width=0.99\textwidth]{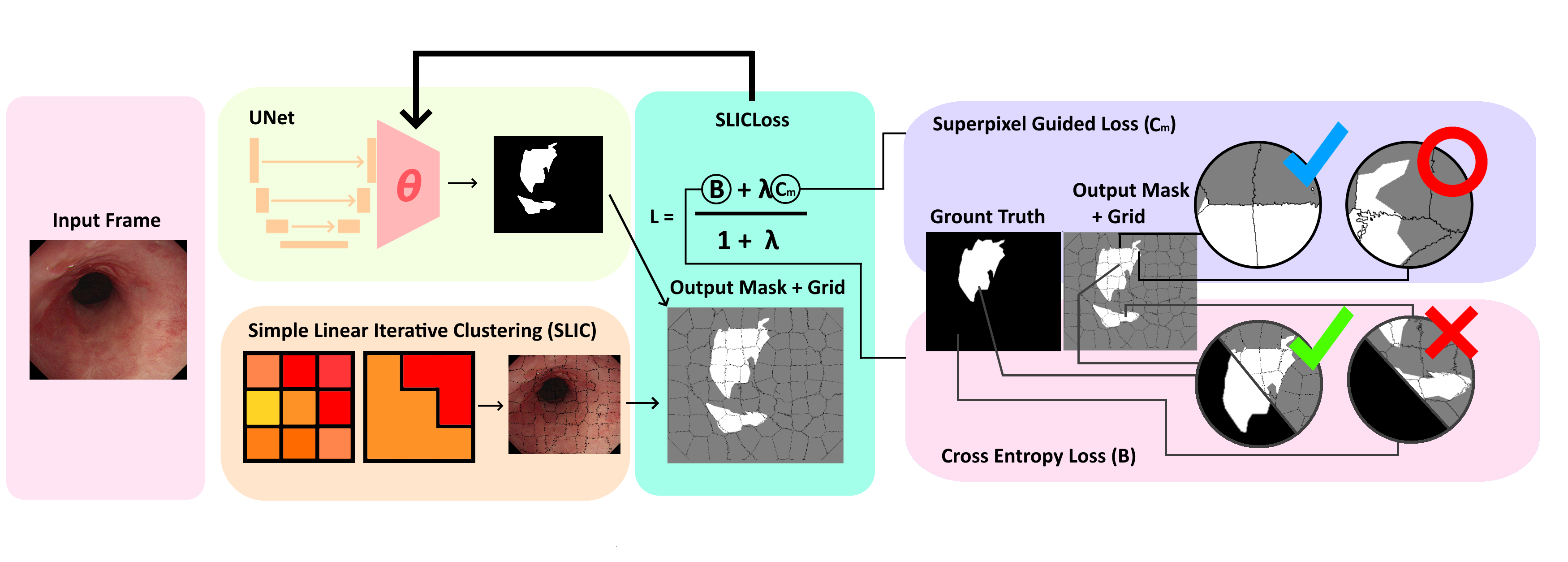}}
\caption{\textbf{A summary of our proposed model.} UNet \cite{ronneberger_u-net_2015} is first trained using only source domain data, with each frame also being used to generate a segmentation output using Simple Linear Iterative Clustering (SLIC) \cite{achanta_slic_2012} which is a $k$-means clustering algorithm. The generated mask from UNet is then combined with the superpixel grid, where two different loss objectives are combined: 1) the superpixel guided loss ($C_m$) that evaluates how closely the mask follows the superpixel boundaries (on the right, a blue check-mark shows a boundary being closely followed, whereas the red circle shows a segmentation that does not follow a boundary), and 2) standard binary cross entropy loss (B), which evaluates the overall accuracy in the prediction (on the right, a green checkmark shows good accuracy, whereas the red cross shows poor accuracy). 
}
\label{fig:summary}
\end{figure*}


Several techniques have been proposed to reduce the effects of domain shift on a model \cite{csurka_unsupervised_2021, tommasi_learning_2016}, which can be divided into either domain adaptation methods (where we have a set of target domains with different data distribution) or domain generalisation methods (where we wish to make our model better with any unseen data). There are many possible directions that one could follow to apply these methods in deep learning: from using conventional algorithms and regularization during training, to modifying the model being used to learn more relevant features or creating additional methods to learn how to model and lessen the impact of domain shift \cite{zhou_domain_2022}.

Domain generalisation methods have shown to be effective in semantic segmentation tasks in areas such as autonomous driving. Thus, in this paper we draw inspiration from this previous work \cite{zhang_transferring_2020} to propose a new loss function that utilizes traditional superpixel segmentation using the Simple Linear Iterative Clustering (SLIC) method, a k-means-based technique \cite{achanta_slic_2012} to enforce a cluster-based consistency during training, generating more geometrically regular predictions that are better translated to other datasets without requiring any changes to the base model. Our contributions herein are two-fold: i) SUPRA (SUPeRpixel Augmented), a framework that generates superpixels and incorporates our loss to a model and ii) SLICLoss, a loss that combines agreement with superpixel boundaries generated by SLIC  and the BCE loss function.

In order to validate our proposal (schematically depicted in fig. \ref{fig:summary}), we make use of an unmodified ('vanilla') UNet model \cite{ronneberger_u-net_2015} and incorporate an additional loss term in the form of the proposed SLICLoss. In our experiments, we compared our method with several baseline models using only the BCE loss on different modalities for BE in EndoUDA. The rationale for this case study is to assess if the proposed loss is capable of improving the network's generalisation capabilities for segmenting images containing a significant domain shift.

The rest of this paper is organized as follows: Section II further discusses the importance of dealing with domain shift in endoscopic tasks, analyzing some recent works in literature. Section III presents the proposed approach and introduces the SLICLoss. In Section IV we discuss the experimental setup, providing training and testing details. Section V presents the results of our experiments in the EndoUDA dataset. Finally, Section VI concludes the article and discusses future work.

\begin{figure*}[htbp]
\centerline{\includegraphics[width=0.8\textwidth]{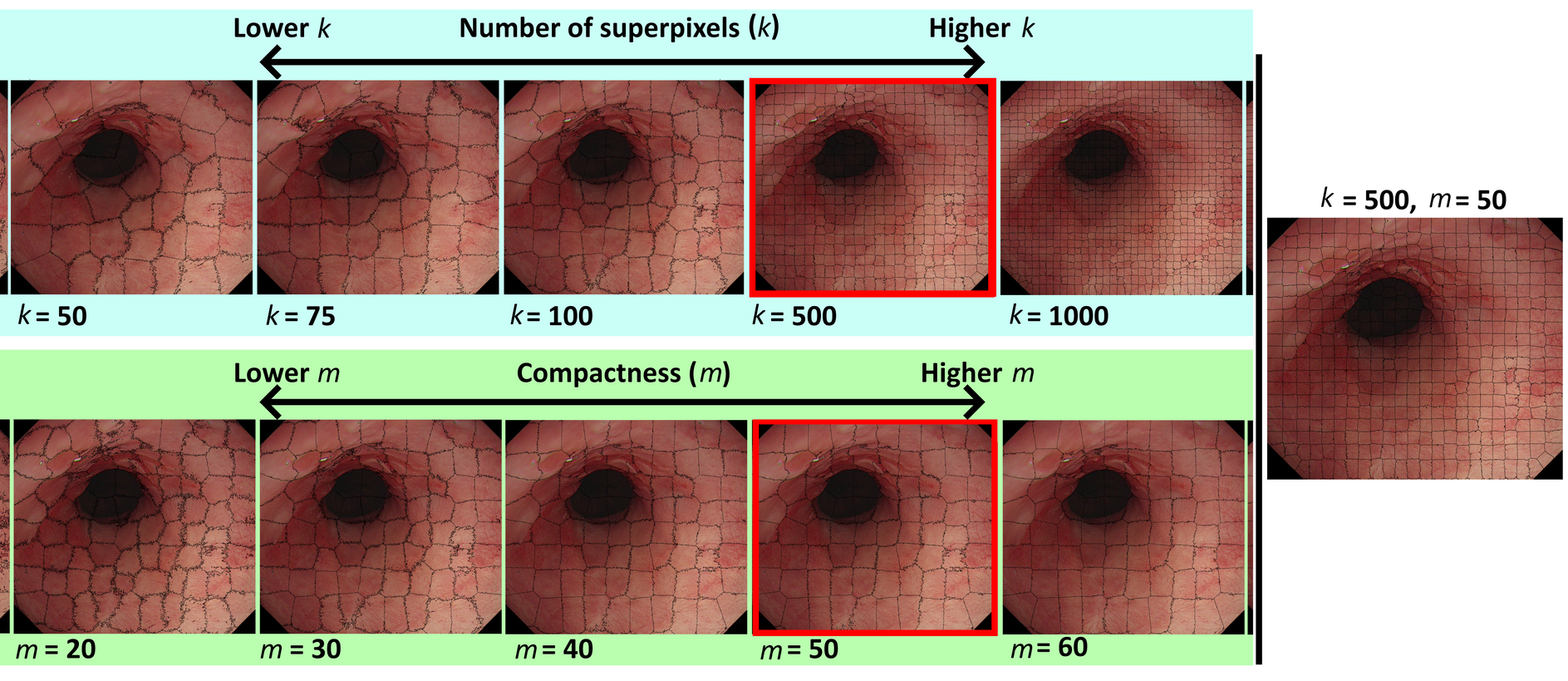}}
\caption{\textbf{Qualitative selection of hyperparameters.} Number of superpixels ($k$, above) shown with a fixed $m$ = 40, and compactness ($m$, below) shown with a fixed $k$ = 100. These were varied from a range of $k$ = 25 to $k$ = 5000 and $m$ = 10 to $m$ = 100 in varying intervals. The values chosen for the final model are shown with a red outline, with an example using the final values appearing on the right side.}
\label{fig:hyperparams}
\end{figure*}

\section{State of the art}
The issue of domain shift is one that has been addressed by many other works in literature. For instance, federated learning has been proposed to leverage many small datasets shared between hospitals to generate a single, large model from them \cite{liu_feddg_2021}. Another approach has been to modify network architectures to learn differently from the same data, and then use those various approaches cooperatively for performing the final prediction  \cite{chen_cooperative_2021}. Other studies, have made use of data augmentation, or have incorporated meta-learning strategies to exploit different characteristics from the training data in a self-supervised manner \cite{zhou_domain_2022}.

Another subset of domain generalisation methods forgo using additional data and instead seek to improve the features learned by a network to make them more relevant to the task. Clustering and patch-based constraints have been used to improve generalisation in road segmentation \cite{zhang_transferring_2020}, and have been shown to reduce the accuracy loss caused by domain shift. The idea behind using constraints is to enforce consistency even if doing so may lead to a decrease in the training accuracy in a particular epoch. This is akin to seeing the domain shift problem as an overfit scenario: the constraints have a regularizing effect that encourages the network to learn more globally applicable features. 

Following the promising results of constraints for domain generalisation, in this work we drew inspiration from superpixel-based methods \cite{zhang_transferring_2020} applied to reduce the domain shift generated by a switch from synthetic to real data. Through the use of SLIC, a superpixel grid is generated without requiring any prior training. In the context of our work, the domain shift we observe arises from different light modalities used in endoscopic interventions. We demonstrate that superpixel patch consistency not only improves results when switching from real to synthetic data as implemented in existing literature, but also increases performance when different lighting modalities are used.

\section{Proposed Approach}

In order to alleviate the problem of domain shift in segmentation tasks for medical imaging,  we propose a framework for a model to favor results that have specific visual properties more globally relevant to the lesions in our data. We identify two qualities that make the use of SLIC superpixels relevant for BE lesions: first, BE heavily alters the color of the esophagus both under WLI and under NBI, suggesting color-based superpixels can provide useful information. Second, lesions in BE are relatively homogeneous, suggesting patch-based consistency is applicable.%
%
%

\begin{table*}[t!]
\begin{center}
\resizebox{\textwidth}{!}{%
\begin{tabular}{lcccc|cc}
\multicolumn{1}{l|}{}                  & \multicolumn{4}{c|}{\textit{Source Domain (White-Light Imaging)}}                                                                    & \multicolumn{2}{c}{\textit{Target Domain (Narrow-Band Imaging)}} \\ \hline
\multicolumn{1}{c|}{\textbf{Network}}  & \multicolumn{1}{l}{\textbf{Validation IoU}} & \multicolumn{1}{l|}{\textbf{Validation Dice}} & \textbf{Test IoU} & \textbf{Test Dice} & \textbf{Test IoU}              & \textbf{Test Dice}              \\ \hline
\multicolumn{1}{l|}{UNet}              & \textbf{0.7602}                             & \multicolumn{1}{c|}{\textbf{0.8454}}          & \textbf{0.6780}   & \textbf{0.7721}    & 0.5359                         & 0.6652                          \\
\multicolumn{1}{l|}{Efficient UNet}    & 0.6228                                      & \multicolumn{1}{c|}{0.7960}                   & 0.5961            & 0.6972             & 0.0545                         & 0.0800                          \\
\multicolumn{1}{l|}{Attention UNet}    & 0.6398                                      & \multicolumn{1}{c|}{0.7557}                   & 0.5737            & 0.6911             & 0.2672                         & 0.3879                          \\ \hline
\multicolumn{1}{l|}{SUPRA-UNet (Ours)} & 0.7587                                      & \multicolumn{1}{c|}{0.8432}                   & 0.6755            & 0.7649             & \textbf{0.6450}                & \textbf{0.7411}                 \\ \hline
\multicolumn{5}{r|}{\textbf{Improvement in Target Domain vs UNet}}                                                                                                                    & +20.35\%                       & +11.41\%                       
\end{tabular}}
\end{center}
\caption{\textbf{Segmentation analysis for the different methods.} SUPRA-UNet: Model trained on proposed loss. UNet: the vanilla UNet was trained only on the BCE loss \cite{ronneberger_u-net_2015}. Efficient UNet: UNet incorporating an EfficientNetb3 backbone \cite{baheti_eff-unet_2020}. Attention UNet: UNet utilizing attention gates \cite{oktay_attention_2018}. The best results are shown with bold formatting. The final row shows the percent difference between our method and the best performing target domain scores for the baseline (obtained by UNet).}
\label{tab1}
\end{table*}

\begin{table*}[t!]
\begin{center}
\resizebox{0.75\textwidth}{!}{%
\begin{tabular}{l|l|c|c|c}
          & \textbf{Hyperparameter} & \textbf{Source Domain IoU} & \textbf{Source Domain Dice} & \textbf{Fixed Parameters} \\ \hline
$\lambda$ & Weight 50\%             & 0.5438                     & 0.6548                      & $k$ = 100,                \\
          & Weight 75\%*            & \textbf{0.7643}            & \textbf{0.8450}             & $m$ = 40                  \\
          & Weight 100\%            & 0.6962                     & 0.7925                      &                           \\ \hline
$k$       & 50 Superpixels          & \textbf{0.7466}            & \textbf{0.8222}             &                           \\
          & 150 Superpixels         & 0.7059                     & 0.7985                      & $\lambda$ = 75\%,       \\
          & 500 Superpixels*        & 0.6881                     & 0.7676                      & $m$ = 40                  \\
          & 1000 Superpixels        & 0.7379                     & 0.7942                      &                           \\ \hline
$m$       & 20 Consistency          & 0.7083                     & 0.8012                      & $\lambda$ = 75\%,         \\
          & 30 Consistency          & \textbf{0.7236}            & \textbf{0.8140}             & $k$ = 100                 \\
          & 50  Consistency*        & 0.4975                     & 0.6151                      &                          
\end{tabular}}
\end{center}
\caption{\textbf{Grid Search.} Experiments performed to observe the effect of different hyperparameters on SUPRA's performance. The validation split from the source domain is used to determine the results.  The models are trained for 15 epochs to generate the preliminary values of these hyperparameters. 
\textbf{$\lambda$:} Weighing factor for the superpixel boundary consistency
$k$: Number of superpixels generated.
$m$: Compactness. 
\textbf{*} Hyperparameter value used for the final model (SUPRA). The best results are shown with bold formatting.}
\label{tab2}
\end{table*}

To achieve our goal of creating a model that is usable in separate modalities without retraining, we propose and test a loss function that penalizes any predictions that generate images that disagree with color differences present in the image. This is done by combining two metrics: Binary Cross-Entropy (BCE) which encourages accurate classification results (as commonly used for binary classification tasks), and a Superpixel Guided Loss, which penalizes the network for generating masks that are not consistent with a coarser superpixel segmentation (see fig. \ref{fig:summary}). To generate the superpixels used for our method, we make use of SLIC, an algorithm that uses color and space distances to generate groups of pixels.

The SLIC superpixel generation algorithm works by using two main parameters \cite{achanta_slic_2012}. The first parameter is $k$, which is the number of superpixels to generate; this parameter enforces the generation of similarly sized regions with spacing $S=\sqrt{N/k}$, where $N$ is the number of pixels in the image. The second parameter of the algorithm is $m$, a constant used to calculate a distance measure used to determine which region a pixel belongs to,
$$D = \sqrt{d_c^2+\left( \frac{d_s}{S}\right)^2 m^2}$$
where $d_c$ is euclidean distance for each color space, and $d_s$ is the euclidean distance between pixels. A higher value of $m$ will encourage compactness, creating regions with a lower area-to-perimeter ratio and more regular shapes. When $m$ is lower, it produces more irregular superpixels that more strictly adhere to areas that present a change in color.

The loss function is constructed from two main elements as can be seen in fig. \ref{fig:summary}: The first is BCE (represented with a \textbf{B} in the fig. \ref{fig:summary}), which evaluates the correctness of the prediction (y') by comparing it to the ground truth (y). The second is a consistency measure identified as Superpixel Guided Loss (represented with a $\mathbf{C_m}$ in fig. \ref{fig:summary}). This measure determines whether the predicted mask produces results that are consistent with the superpixel segmentation generated by SLIC operating on the input frame (x). 

The $C_m$ loss works by determining how much of a superpixel's area is occupied by a single class. This is done by taking the difference of each class' occupied area within a superpixel, and comparing it with a threshold. Any superpixel that is occupied by less than the threshold is considered to be inconsistent with the expected boundary
(red circle in fig. \ref{fig:summary}). Every superpixel is evaluated in this manner, with the inconsistencies summed and averaged to produce a final loss metric.

To combine BCE and $C_m$, our loss is multiplied by a weighing factor ($\lambda$), and both losses are added and scaled by 1+$\lambda$. The result is then used as the loss for the network.

\begin{equation}
    L(x,y,y')=\frac{BCE(y,y')+\lambda C_m(x,y')}{1+\lambda}
    \label{eq:corr}
\end{equation}
As can be seen in eq. \ref{eq:corr}, the weighing factor ($\lambda$) can be adjusted to favor more accurate results, or more superpixel consistent results. A higher $\lambda$ will strengthen the effect of the superpixels, at the cost of disregarding overall accuracy. $\lambda$, $k$, $m$, and the threshold for $C_m$ are hyperparameters that must be tuned in accordance to properties in the source domain. This tuning is explained in section \ref{sec:hyperparams}.

\section{Experimental Design}
\subsection{Data Partitioning}

\begin{table*}[t!]
\begin{center}
\resizebox{0.75\textwidth}{!}{%
\begin{tabular}{l|l|c|c|l}
          & \textbf{Hyperparameter} & \textbf{Target Domain IoU} & \textbf{Target Domain Dice} & \textbf{Fixed Parameters} \\ \hline
$\lambda$ & Weight 50\%             & \textbf{0.5944}            & \textbf{0.7221}             & $k$ = 100,                 \\
          & Weight 75\%*            & 0.5865                     & 0.7217                      & $m$ = 40                   \\
          & Weight 100\%            & 0.4522                     & 0.6089                      &                          \\ \hline
$k$       & 50 Superpixels          & 0.4937                     & 0.6362                      &                          \\
          & 150 Superpixels         & 0.3738                     & 0.5228                      & $\lambda$ = 75\%,        \\
          & 500 Superpixels*        & \textbf{0.6429}            & \textbf{0.7679}             & $m$ = 40                   \\
          & 1000 Superpixels        & 0.4135                     & 0.5619                      &                          \\ \hline
$m$       & 20 Consistency          & 0.3650                     & 0.5153                      & $\lambda$ = 75\%,        \\
          & 30 Consistency          & 0.3872                     & 0.4763                      & $k$ = 100                  \\
          & 50  Consistency*        & \textbf{0.5919}            & \textbf{0.7212}             &                         
\end{tabular}}
\end{center}
\caption{ \textbf{Effect of Hyperparameters on Target Domain.} Target domain scores of the models trained using the hyperparameter grid search as shown in table \ref{tab2}. Although not used to select the final parameters, these results show the shift between the source domain validation and target domain test performance. Instead of using the grid search, we opted to use a qualitative assessment of the generated superpixel boundaries to pick the best values for $k$ and $m$. 
\textbf{$\lambda$:} Weighing factor for the superpixel boundary consistency
$k$: Number of superpixels generated.
$m$: Compactness. 
\textbf{*} Hyperparameter value used for the final model (SUPRA). The best results are shown with bold formatting.}
\label{tab3}
\end{table*}

 The dataset used to perform our experiments is the EndoUDA \cite{celik_endouda_2021, ali_deep_2021, borgli_hyperkvasir_2020}, which contains endoscopic images from two medical tasks (binary segmentation for BE and polyps). EndoUDA is composed of 799 images for BE, with 284 in NBI modality and 515 in (WLI) modality. It also contains 1042 images for polyps, with 42 in NBI modality and 1000 in WLI modality (see fig. \ref{fig:endouda}). We test our method only for the BE split of the dataset, as SLIC is not adequate for the topologically difficult characteristics of polyps. For such tasks, other superpixel generation methods may be more appropriate.We used a 80/10/10 split on the WLI frames for training, validation, and testing (respectively) and used every frame available for NBI during testing (284 frames).

\subsection{Baselines}
In addition to our method, we train three other models: Efficient UNet \cite{baheti_eff-unet_2020}, Attention UNet \cite{oktay_attention_2018}, and a vanilla UNet \cite{ronneberger_u-net_2015}. These serve to observe both the performance of other models in the dataset being tested, and to confirm the presence of domain shift between the two lighting modalities: If the source domain testing presents higher accuracy than our target domain, then domain shift exists and it is adversely affecting the models' ability to generalise to NBI.

\begin{figure*}[htbp]
\centerline{\includegraphics[width=0.62\textwidth]{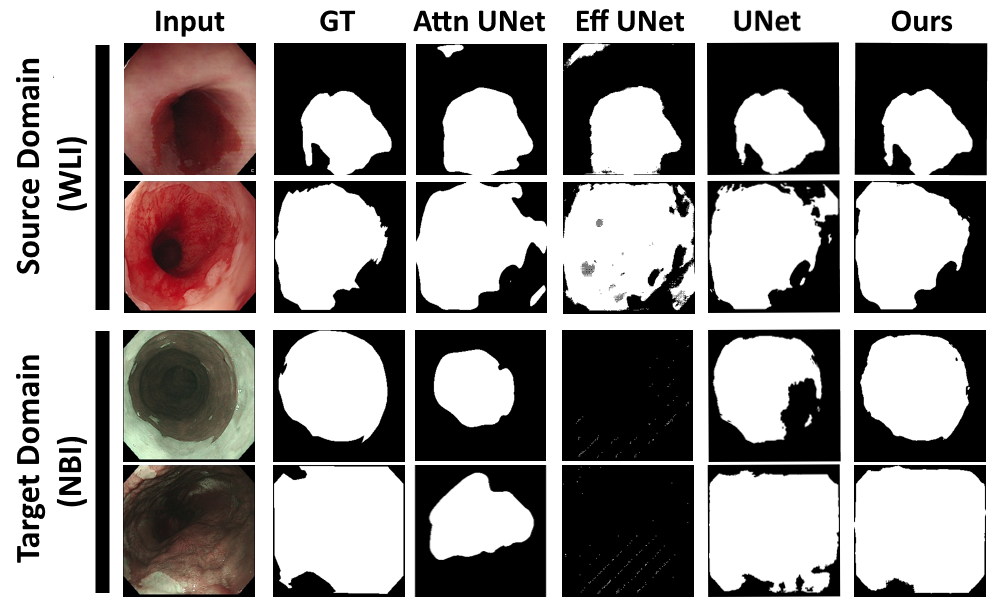}}
\caption{\textbf{Qualitative comparison.} We include frames from tested models, comparing between SUPRA-UNet (Ours), vanilla UNet, Attention UNet (Attn UNet), and Efficient UNet (Eff UNet). For all splits, SUPRA-UNet has competitive results on the source domain but out-performs other baseline methods on target domain samples.}
\label{fig:qual}
\end{figure*}

\subsection{Training}

All models were trained using a validation loss early stop with patience of 15 epochs, saving the best validation loss weights. In table \ref{tab1}, we present the results of all the tested models. All were trained using Tensorflow 2.8.1 on a RTX 2060 GPU  using CUDA 11.2, with a learning rate of $1e^{-04}$, using an Adam optimizer. The batch size was set to 1 for all runs due to GPU memory constraints. Images were resized from 1164 x 1030 to 256 x 256 using bilinear interpolation for all models. Geometric data augmentations were used in the training set, using the default Tensorflow generators. These include horizontal mirroring, rotation, width and height shift, shearing, and zooming. The maximum strength for these transformations were set to 5\% where applicable, except for rotation which was set to 20\%.

\subsection{Hyperparamater tuning}
\label{sec:hyperparams}
SLIC-produced superpixels are highly dependent on the parameters used. To ensure the superpixels were being adequately generated, values for $\lambda$, $k$ and $m$ were adjusted through a qualitative observation of the best superpixel grids generated on a small subset of the training images (see fig. \ref{fig:hyperparams} for an example). 

A first approach was to utilize a grid search (see table \ref{tab2}), using a preliminary training of 15 epochs to test $\lambda$, $k$, and $m$ at different combination. However, the first iterations of this model showed poor results after the parameter selection. Further diagnosing revealed that the parameters $k$ and $m$ did not translate from validation to test domain with the metrics used. This is further evident by observing the test performance of our grid search hyperparameters (see table \ref{tab3}), but such results are unusable for parameter selection as they run the risk of introducing target domain information to the network. Instead, we opted to perform a qualitative selection of the best parameters.

Our alternative approach involved the qualitative observation of superpixel qualities as presented in fig. \ref{fig:hyperparams}. This involved varying $k$ with $m$ fixed, and viceversa, to observe the change in two image qualities: The adherence to the observed boundary of different features of the lesion (that is, if a superpixel segmentation correctly separated differently colored areas of the image), and the smoothness of the results (the amount of boundaries that followed noise instead of the ground truth segmentation). Varying the value of $k$ caused two visible effects: At very low values ($k$ = 50 and below), there would be significant undersegmentation causing lesions to be grouped together with healthy tissue. At very high values ($k$ = 1000 and above), all noise is segmented in individual superpixels -- this could cause the network to learn incorrect information that does not relate to the injury.

Varying $m$ was more straightforward: a smaller $m$ is directly related to better boundary adherence (better at differentiating colors), but the boundary becomes noisier.  A higher $m$ increases smoothness, but also decreased boundary adherence.  With this criteria in mind, we found that the values $m$ = 50, and $k$ = 500 are a good compromise. A further increase in $m$ leads to superpixels ignoring color boundaries, while decreases suffered from too much irregularity in segmentation that strayed from the ground truth masks. Similarly, increasing $k$ leads to noise being individually grouped into superpixels. Decreases from $k$ = 500 led to some undersegmentation. Since $\lambda$ cannot be tested qualitatively in the same manner, we opted to select it based on the grid search. Following the validation results, $\lambda$ was set to a value of $\lambda$ = 75\%. 

For $C_m$, the threshold for the expected occupation was set to 80\%, as this was the observed minimum percentage of occupation in the GT masks used for the hyperparameter tuning. Our final model was trained with early stopping using the final configuration of: $k$= 500 superpixels, $m$ = 50, $\lambda$ = 75\%, and $C_m$ threshold = 80\%.

%

\begin{figure*}
    \centering
    \begin{subfigure}{0.37\textwidth}
        \includegraphics[width=\textwidth]{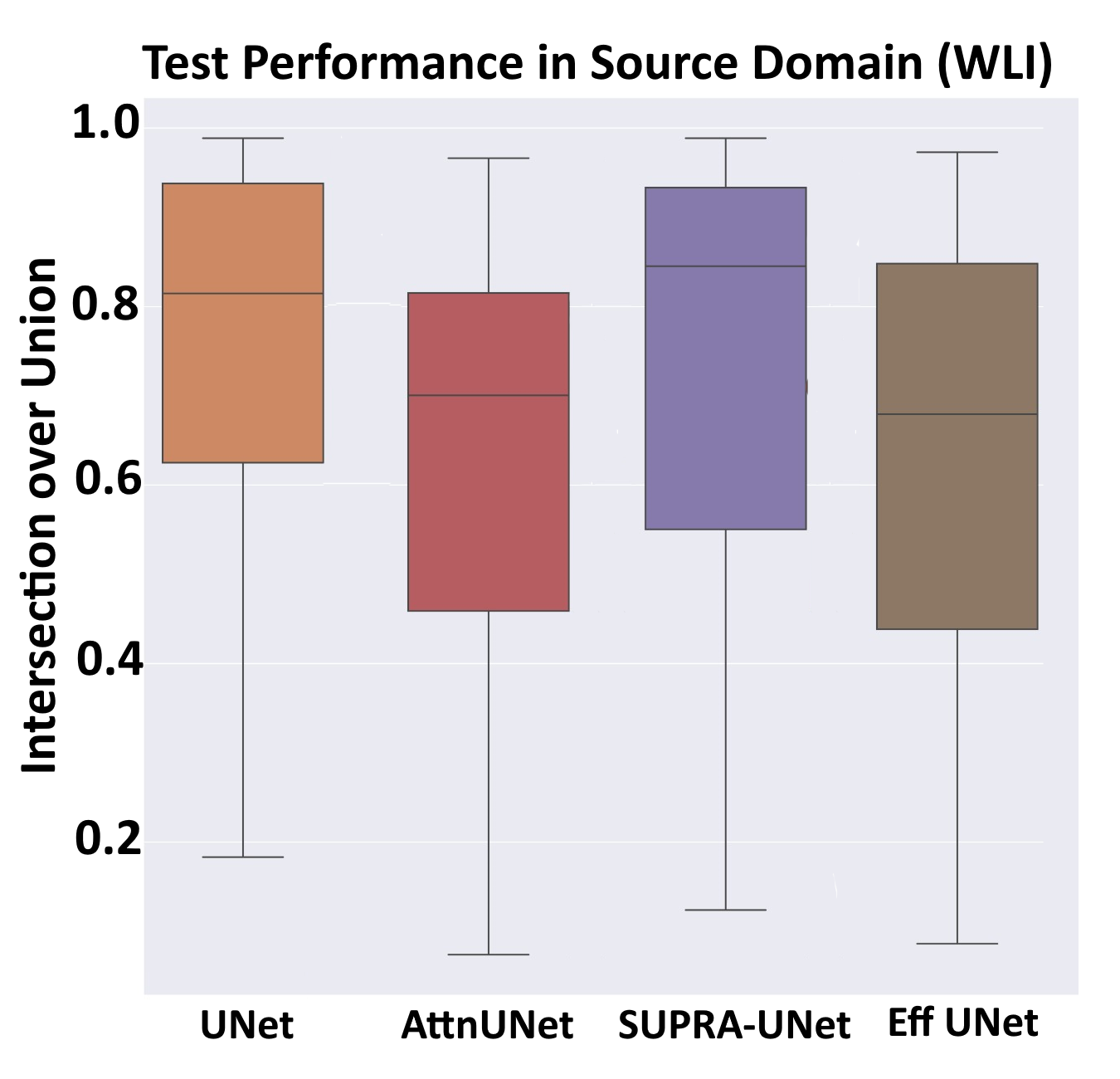}
        \caption{}%
        \label{sfig:iid_box}
    \end{subfigure}\hspace{0.1\textwidth}
    \begin{subfigure}{0.37\textwidth}
        \includegraphics[width=\textwidth]{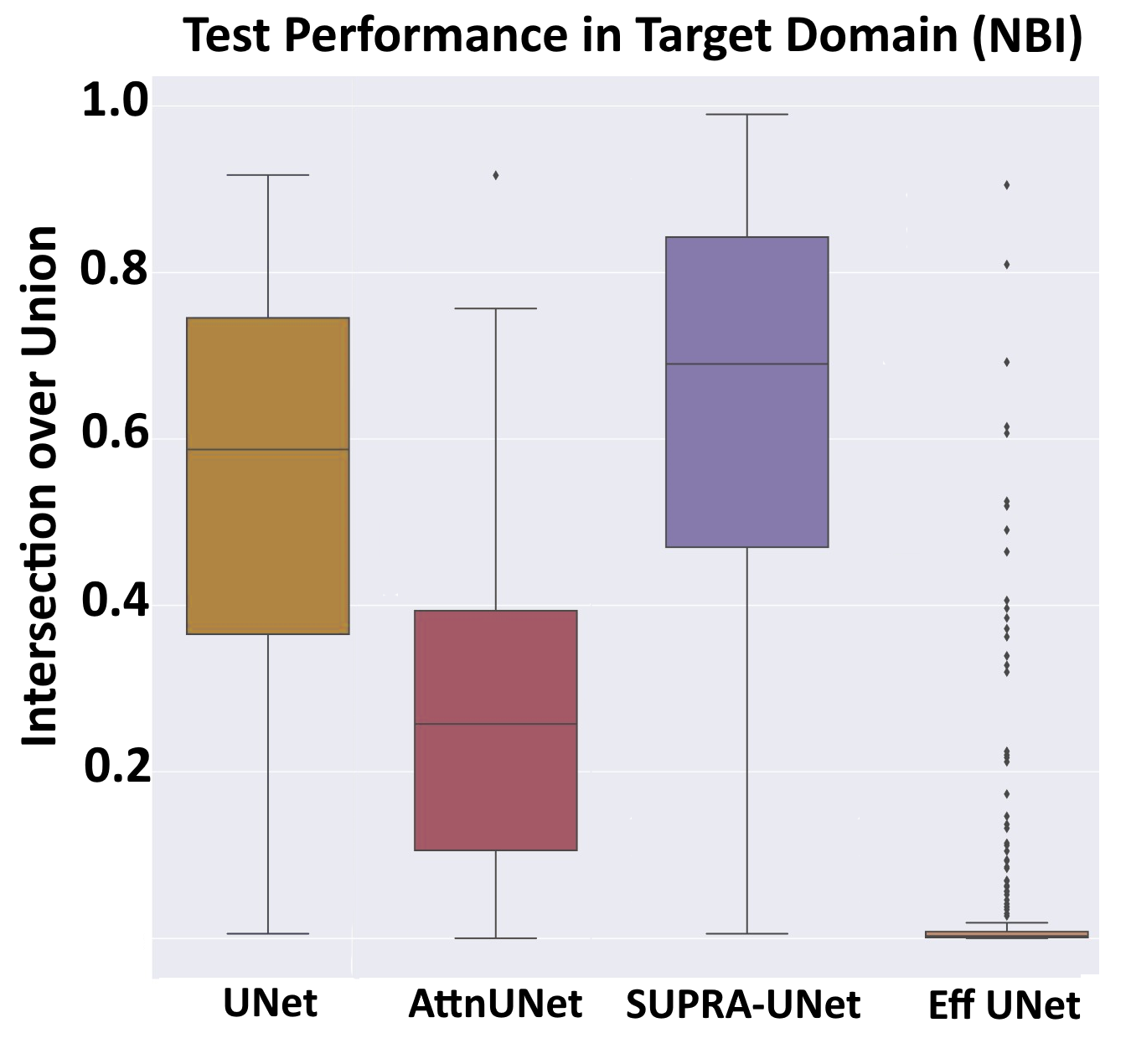}
        \caption{}%
        \label{sfig:ood_box}
    \end{subfigure}
    \caption{\textbf{Box plots of tested models.} Attn UNet: Attention UNet. SUPRA-UNet: Our method. Eff UNet: Efficient UNet. For the source domain, performance was similar across all models, with a relatively wide range in results. The best performing models were UNet and SUPRA-UNet, with UNet achieving a smaller median but with less variation in the results, whereas SUPRA-UNet had a higher median but more variable results. For the target domain, SUPRA-UNet higher median and less variation.}
    \label{fig:boxplotsi}
\end{figure*}

\section{Results}
The results for the experiments are summarized in table \ref{tab1}. We can observe that there is a considerable drop in segmentation performance when using BCE loss from the validation set (Source Domain, SD) to the test set (Target Domain, TD) in all models, suggesting the existence of severe domain shift. This is particularly notable in Attention and Efficient UNet. UNet loses around 0.14 IoU when switching between lighting modalities, having performed the best out of all control models used.

Notably, vanilla UNet achieved the best in-domain results when compared to other unmodified models. Attention mechanisms are best when there are areas in the image where a network should focus on to find the most relevant features to achieve segmentation, and they are likely superfluous when the only region to focus on is the lesion itself. Efficient UNet, likewise, achieves lower performance: Here, the choice of backbone is a possible culprit, as it was observed that the early stop for Efficient UNet was sooner than those for the other networks (about 30 epochs vs about 50 epochs). This could signify that the backbone chosen (EfficientNetB3) is too complex for the relatively straightforward task of lesion segmentation, but further experimentation is required to confirm this suspicion.

When compared to vanilla UNet, SUPRA-UNet quantitatively performed nearly equally as good in the source domain, with less than 0.01 difference in both Dice and IoU. This, however, is not the same in the target domain: SUPRA-UNet handily beats all tested models with an increase of 0.11 IoU against UNet's result, and a degradation of only 0.03 IoU when switching lighting modalities. Here, we notice that the expected regularization is happening: given a slight reduction in SD performance, we have significant increases in the TD performance. We can intuitively think of this change as the superpixels being unable to fully describe every lesion with full accuracy, so some detail is lost when the network is forced to heed the boundaries given by them. This comes at the benefit that the superpixels, however, can and do better describe the lesions present in NBI when compared to an unaugmented network.

The results of this improved representation can be observed in fig. \ref{fig:qual}, which shows the predicted masks from all tested networks. The boundaries generated by SUPRA-UNet are more accurate to the color differences in the images. Likewise our method is less prone to generate patches: it creates masks devoid of internal discontinuities that favor homogeneous results. 

Since SLIC uses color-space features, our boundaries follow color differences more closely. This is in contrast to the other models: The features learned, while accurate in the case of UNet, are based around a more complex understanding of the image. While these features lend themselves to marginally better performance in the SD, they generalize poorly to the TD. Furthermore, the patch-based consistency that SLICLoss incorporates discourages a common cause for error both in- and out-of-domain images experience: Masks with noise that cause highly heterogeneous predictions (areas with both lesion and background classes).

Boxplots were generated for a statistical analysis, and can be seen in fig. \ref{fig:boxplotsi}. These take into account the 51 test data points for WLI (10\% of the total 515 images), and the 284 frames for NBI. For in-domain data (fig. \ref{sfig:iid_box}), minimums were between 0.1 and 0.2 IoU for all models, with performance close to 0.7-0.8 for their medians, similar to the mean observed in table \ref{tab1}. UNet and SUPRA-UNet perform similarly, with SUPRA-UNet producing more variable results but with a better mean.

For out-of-domain data (fig. \ref{sfig:ood_box}), minimums were 0 for all models, with some images still being too different from the source domain to be accurately segmented. Efficient UNet and Attention UNet had very poor performances, with Efficient UNet in particular only having a few outliers that achieved acceptable performance. SUPRA-UNet and UNet performed better, although both presented high amount of variability. For UNet, the explanation is simple: Some examples were similar enough to the target domain that the network could predict in the same way it did in the source domain. For SUPRA-UNet, the variability observed in the target domain is related to the fact that although our method does improve domain shift, it does not eliminate it: It still experiences a difference of roughly 0.15 IoU compared to its source domain results. Taking this factor into account, our method achieves a higher overall performance, both in the maximums and with most of the dataset achieving higher IoU than the other models.

\section{Conclusions}
SUPRA-UNet yielded a significant improvement in our BE target dataset, with a 0.11 increase in IoU when compared to using only BCE as loss for training using white-light imaging (WLI) as the source domain and predicting in narrow-band imaging (NBI) as the target domain. This is done without any degradation in source domain performance: We find that it performs statistically very close to UNet. Even with a quantitative similar performance in the SD set, it was still able to obtain better qualitative results in both domains. In the future, other superpixel generation methods can be tested to incorporate more complex topological information. Such methods may be able to provide the same improvement in tasks such as polyp segmentation by incorporation structural details in either the loss function or in the superpixel generation process. 

\section*{Acknowledgments}

The authors wish to thank the AI Hub and the CIIOT at Tecnologico de Monterrey for their support for carrying part of the experiments reported in this paper in their NVIDIA DGX computer. We also wish thank CONACYT for the master scholarship for Rafael Martinez Garcia Peña at Tecnologico de Monterrey.

{\small
\bibliographystyle{ieee_fullname}
\bibliography{ISBI2023}
}

\end{document}